\documentclass[11pt,a4paper]{article}
\usepackage{authblk}
\usepackage[hyperref]{acl2020}
\usepackage{times}
\usepackage{latexsym}

\usepackage{array}
\newcolumntype{P}[1]{>{\centering\arraybackslash}p{#1}}
\usepackage{microtype}

\aclfinalcopy 

\title{Experimental Evaluation of Deep Learning models for Marathi Text Classification}

\author[Atharva Kulkarni et al.]
       {Atharva Kulkarni$^1$, Meet Mandhane$^1$, Manali Likhitkar$^1$,
       Gayatri Kshirsagar$^1$, 
       \\
        Jayashree Jagdale$^1$ and Raviraj Joshi$^2$
       \\
       $^1$Pune Institute of Computer Technology, Pune\\
       $^2$Indian Institute of Technology Madras, Chennai\\
       \{k.atharva4899,meetmandhanemnm,manalil1806,gayatrimohan7\}@gmail.com\\ 
       \{jayashree.jagdale,ravirajoshi\}@gmail.com\\ 
       }
\date{}

\begin{document}
\maketitle
\begin{abstract}
The Marathi language is one of the prominent languages used in India. It is predominantly spoken by the people of Maharashtra. Over the past decade, the usage of language on online platforms has tremendously increased. However, research on Natural Language Processing (NLP) approaches for Marathi text has not received much attention. Marathi is a morphologically rich language and uses a variant of the Devanagari script in the written form.\\
This works aims to provide a comprehensive overview of available resources and models for Marathi text classification. We evaluate CNN, LSTM, ULMFiT, and BERT based models on two publicly available Marathi text classification datasets and present a comparative analysis. The pre-trained Marathi fast text word embeddings by Facebook and IndicNLP are used in conjunction with word-based models. We show that basic single layer models based on CNN and LSTM coupled with FastText embeddings perform on par with the BERT based models on the available datasets. We hope our paper aids focused research and experiments in the area of Marathi NLP.
\end{abstract}

\section{Introduction}

\label{sec:1}
The Marathi language is spoken by more than 83 million people in India. In terms of the number of speakers, it ranks third in India after Hindi and Bengali. It is native to the state of Maharashtra and also spoken in Goa and some regions of western India. Despite leading in education and economy, NLP research in Marathi has not received much attention in Maharashtra. In contrast research in Hindi has been much more significant \cite{arora2013sentiment,akhtar2016aspect,joshi2016towards,joshi2019deep,patra2015shared} followed by Bengali \cite{patra2018sentiment,al2017sentiment,pal2015automatic,sarkar2017sentiment}. Recently, to enable cross-lingual NLP, the focus has shifted to building single multilingual models instead of models for individual language \cite{pires2019multilingual, conneau2019cross}. \\

English is the most widely used language globally, due to which the majority of the NLP research is done in English. There is a large scope for research in other regional languages in India. These regional languages are a low resource and morphologically rich, which is the main reason for limiting research \cite{patil2017mars}. Morphologically rich languages have greater complexity in their grammar as well as sentence structure. Grammatical relations like subject, predicate, object, etc., are indicated by changes to the words. Also, the structure of the sentence in Marathi can change, without it affecting the meaning. For example, "To khelayla yenar nahi", "Khelayla nahi yenar to", "To nahi yenar khelayla". An absence of large annotated datasets is a major issue due to which new methods cannot be properly tested and documented, thus hampering research. In this work, we are concerned with Marathi text classification. We evaluate different deep learning models for Marathi text and provide a comprehensive review of publicly available models and datasets.\\

Text classification is the process of categorizing the text into different classes, grouped according to content \cite{kowsari2019text}. It has been used in a variety of applications from optimizing searches in search engines to analyzing customer needs in businesses. With the increase in Marathi textual content on online platforms, it becomes important to build text processing systems for the Marathi language. The text classification module requires the application of various pre-processing techniques to the text before running the classification model. These tasks involve steps viz tokenizing, stop word removal, and stemming the words to their root form. Tokenization is a way of separating a piece of text into smaller units called tokens. The tokens can be either word, characters, or subwords. Tokens and punctuations that do not contribute to classification are removed using stopword removal techniques. Stemming is the process of reducing a word to its root form. These root forms of words then represent the sentence of the document to which they belong and are passed on to classifiers. In this work, we are not concerned about stemming and stop word removal. More recent techniques based on sub-words and neural networks implicitly mitigate problems associated with morphological variations and stop words to a large extent \cite{joulin2016bag}.\\

This paper explores and summarizes the efficiency of Convolutional Neural Network (CNN), Long Short Term Memory (LSTM), and Transformer based approaches on two datasets. The two datasets are contrasting in terms of length of records, grammatical complexity of sentences, and the vocabulary itself. We also evaluate the effect of using pre-trained FastText word embeddings and explicit sub-word embeddings on the aforementioned architectures. Finally, we evaluate the pre-trained language models such as Universal Language Model Fine-tuning (ULMFiT) and multilingual variations of the Bidirectional Encoder Representations from Transformers (BERT), mBERT, and IndicBERT for the task of Marathi text classification \cite{howard2018universal, devlin2018bert}. We use the publicly available ULMFiT and BERT based models and re-run the experiments on the classification datasets. The evaluation of other models and the effect of using pre-trained word embeddings on them is specific to this work and not covered in previous literature. The main contributions of this work are
\begin{itemize}
    \item We provide an overview of publicly available classification datasets, monolingual corpus, and deep learning models useful for the Marathi language. We emphasize that Marathi is truly a very low resource language and even lacks a simple sentiment classification dataset.
    \item We evaluate the effectiveness of publicly available FastText word embeddings and sub-word tokenizers for Marathi text classification. We present a comparative analysis of CNN, LSTM, and Transformer based models. The analysis shows that simple CNN and LSTM based models along with FastText embeddings performs as good as currently available pre-trained multilingual BERT based models.
\end{itemize}

\section{Related Work}

\label{sec:2}
There has been very little research done on Marathi NLP. Recently, \cite{kakwani2020indicnlpsuite} introduced NLP resources for 11 major Indian languages. It had Marathi as one of the languages. They collected sentence-level monolingual corpora for all the languages from web sources. The monolingual corpus contains 8.8 billion tokens across multiple languages. This corpus was used to pre-train word embeddings and multi-lingual language models. The pre-trained models are based on the compact ALBERT model termed IndicBERT. The FastText embeddings were trained as it is better at handling morphological variants. The pre-trained models and embeddings were evaluated on text classification and generation tasks. In most of their experimentations, IndicBERT models have outperformed XLM-R and mBERT models. They have also created other datasets for various NLP tasks like Article Genre Classification, Headline Prediction, etc. This work is referred to as IndicNLP throughout the paper.\\

An NLP library for Indian languages iNLTK is presented in \cite{arora2020inltk}. It consists of pre-trained models, support for word embeddings, textual similarity, and other important components of NLP for 13 Indian languages including Marathi. They evaluate pre-trained models like ULMFiT and TransformerXL. The ULMFiT model and other pre-trained models are shown to perform well on small datasets as compared to raw models. Work is being done to expand the iNLTK support to other languages like Telugu, Maithili, and some code mixed languages.\\

Previously, \cite{bolaj2016text} presented supervised learning methods and ontology-based Marathi text classification. Marathi text documents were mapped to the output labels like festivals, sports, tourism, literature, etc. The steps proposed for predicting the labels were preprocessing, feature extraction and finally applying supervised learning methods. Methods based on Label Induction Clustering Algorithm (LINGO) to categorize the Marathi documents were explored in \cite{vispute2013automatic, patil2015automatic}. A small custom data-set containing 100-200 documents was used for classification in the respective work.\\

\section{Data Resources}
\label{sec:3}
\subsection{Datasets}
This section summarizes publicly available classification datasets used for experimentation.

\paragraph{IndicNLP News Article Dataset: }This dataset consists of news articles in Marathi categorized into 3 classes viz. sports, entertainment, and lifestyle \cite{kakwani2020indicnlpsuite}. The dataset contains 4779 records with predefined splits of the train, test, and validation sets. They contain 3823, 479, and 477 records respectively. The average length of a record is approximately 250 words.

\paragraph{iNLTK Headline Dataset: }This dataset consists of Marathi news article headlines of 3 different classes viz. entertainment, sports, state \cite{arora2020inltk}. It consists of 12092 records. 
The dataset made available under the IndicNLP catalog consists of 9672 train, 1210 test, and 1210 validation samples. The average length of a record is 7 words.

\subsection{Embedding}
The input to the classification models can be a word, sub-word, or character embeddings. In this work, we have experimented with word and sub-word embeddings. For word embeddings, random initialization and FastText initialization are explored. Whereas for sub-word embeddings we train a unigram based sentence piece tokenizer and randomly initialize the sub-word embeddings \cite{kudo2018sentencepiece}. A vocab size of 12k is used for sub-word tokens. There are two versions of pre-trained Marathi FastText embedding models available trained by Facebook and IndicNLP. We use both of these in static and trainable mode. In static mode, the word-embedding layer is frozen and in trainable mode, the embedding layer is kept trainable.

\begin{table*}
\begin{center}
\begin{tabular}{{p{1.5cm}p{4.5cm}P{2cm}P{2cm}}}
\hline 
\textbf{Models} & \textbf{Variants} & \textbf{News Articles} & \textbf{News Headlines}  \\
\hline
CNN & random-word & 98.95 & 89.01 \\
& random-subword & 98.53 & 88.18 \\
& FBFastText-Trainable & 98.74 & 91.49 \\
& FBFastText-Static & 96.65 & 90.50 \\
& IndicNLP FastText-Trainable & 99.16 & \textbf{94.88}\\
& IndicNLP FastText-Static & \textbf{99.37} & 94.13\\
\hline
LSTM & random-word & 98.74 & 88.51 \\
& random-subword & 97.69 & 86.36 \\
& FBFastText-Trainable & 98.95 & 91.49 \\
& FBFastText-Static & 88.22 & 89.75 \\
& IndicNLP FastText-Trainable & \textbf{99.16} & \textbf{94.79}\\
& IndicNLP FastText-Static & 98.54 & 94.55\\
\hline
BiLSTM & random-word & 98.32 &  89.09 \\
& random-subword & 96.86 & 86.78 \\
& FBFastText-Trainable & 97.70 & 91.24 \\
& FBFastText-Static & 94.98 & 89.83 \\
& IndicNLP FastText-Trainable & \textbf{99.16} & \textbf{94.63} \\
& IndicNLP FastText-Static & 99.16 & 94.13 \\
\hline
ULMFiT & (iNLTK) & \textbf{99.37} & \textbf{92.39} \\
\hline
BERT & mBERT & 97.48 & 90.70 \\
& IndicBERT(INLP) & \textbf{99.40} & \textbf{94.50} \\
\hline
\end{tabular}

\caption{\label{tab:1} Classification accuracies over different architectures }
\end{center}
\end{table*}

\subsection{Monolingual Corpora}
Although we have not explicitly used monolingual corpus in this work, we list the publicly available Marathi monolingual corpus for the sake of completeness. These individual corpora were used to pre-train FastText word embeddings, ULMFiT, and BERT based models by the respective authors.

\paragraph{Wikipedia text corpus: } The Marathi Wikipedia article monolingual dataset consists of 85k cleaned articles. This is a small corpus which consists of comparatively fewer tokens.

\paragraph{CC-100 Monolingual Dataset: } The dataset is a huge collection of crawled websites for 100+ languages \cite{wenzek2019ccnet}. It was created by processing January-December 2018 Commoncrawl snapshots. However, for Marathi as well as most other Indian languages, the dataset consists of just about 50 million tokens each.

\paragraph{OSCAR: } Open Super-large Crawled ALMAnaCH coRpus (OSCAR) is obtained by filtering and language classifying the Common Crawl corpus \cite{suarez2019asynchronous}. After de-duplifying all words, the size of the Marathi corpus comes up to 82 million tokens.

\paragraph{IndicNLP Corpus: } This is a multi-domain corpus that spans over 10 Indian languages and contains over 2.7 billion tokens. For Marathi, the dataset consists of 9.9 million sentences, 142 million tokens.

\section{Architecture}
Traditionally, LSTM based models are used for NLP activities. However, recent experiments have shown that CNN based models give encouraging results \cite{joshi2019deep,kim2014convolutional}. We have explored a variety of models, which give insights into their applications according to the data used.

\paragraph{CNN: }An embedding layer converts inputs to word embeddings of length 300. This sequence of word embeddings is passed on to a Conv1D layer with 300 filters and kernel size 3. This layer is followed by a GlobalMaxPool layer and a dense layer of size 100. A final dense layer of size equal to the number of target labels completes the model.

\paragraph{LSTM + GlobalMaxPool: }
An embedding layer outputs embeddings of length 300. This is followed by an LSTM layer with cell size 300 and then a GlobalMaxPool layer. The output is then fed to a dense layer of size 100, and finally a dense layer of size equal to the total number of target classes.

\paragraph{BiLSTM + GlobalMaxPool: }
The word vectors from the embedding layer are passed onto a Bidirectional LSTM layer with cell size 300. The output of the Bi-LSTM layer is max pooled over time. This is followed by a dense layer of size 100 and a final dense layer.

\paragraph{ULMFiT: } 
Universal Language Model Fine-tuning for Text classification uses transfer learning and can be used for various NLP activities. The language model pretraining allows it to work well on small datasets. The pre-trained models made publicly available by iNLTK is finetuned for our target dataset.

\paragraph{BERT: } 
Various transformer-based BERT language models pre-trained from text corpus is publicly available. We tested out some of the versions and fine-tuned it for the Marathi text classification task, viz 
\begin{itemize}
    \item Multilingual-BERT
    \item Indic-BERT trained by IndicNLP
\end{itemize}
For each of these models, we experimented with two variations: a CLS-based architecture for sequence classification; and fine-tuning the final layers of the model and passing them to a classifier layer.

\section{Results}

We evaluated different CNN, LSTM, BiLSTM based models along with language models such as ULMFiT and BERT on two datasets viz. Marathi News Headline Dataset and Marathi News Articles Dataset. The two input representations based on words and sub-words are used. For word-based representations, we compare random initialization and FastText based initialization. The FastText based initialization is then used in trainable and non-trainable mode. Table \ref{tab:1} shows a comparison of various approaches. The variant column indicates the variation used for input representation. The random initialization of embeddings is indicated as random-word and random-subword. The original FastText embedding released by Facebook is termed as FBFastText. The version released by IndicNLP is termed IndicNLP FastText. The suffix static indicates an un-trainable embedding layer while trainable indicates a trainable one.\\

For both the datasets, the IndicBERT, and the models that use IndicNLP FastText Embeddings gives the best results. For the articles dataset, the increase in accuracy after using IndicNLP embeddings over a simple model is small, but the increase is significant for the headline dataset. Also, the FBFastText embeddings do not match up to the results of the IndicNLP embeddings. This is in line with the observations in the original work. The CNN based models have a slight edge over LSTM and BiLSTM based models. The ULMFiT model performs better than models using random and FBFastText word embeddings.\\

Finally, for the BERT-based models, we experimented with fine-tuning the final layers of the BERT model, passing them onto a BiLSTM model and only extracting the sentence embeddings ie CLS tokens, and using them for classification. Both approaches give us almost similar results. The IndicBERT works best for both datasets. The results for mBERT and models using FBFastText are in the same range. Moreover, the IndicBERT and models using IndicNLP FastText also perform in the same range but roughly 4\% better than mBERT in absolute terms. The ULMFiT is somewhere between mBERT and IndicBERT. Overall, models utilizing IndicNLP FastText embeddings perform competitively with IndicBERT. The higher accuracy of these models can be attributed to the fact that a major chunk of data used to pre-train them came from news sources. The target datasets also come from the news domain. \\

The CNN + IndicNLP FastText performs best on the Headlines dataset leaving behind IndicBERT by a small margin. However, IndicBERT is multilingual in nature and serves more applications. The results in Table 1 show that for applications only concerned about Marathi text, basic models will be more preferable.\\

\section{Conclusion}

In this paper, we have experimented with various deep learning models and evaluated them for the task of Marathi text classification. We present a comparative analysis of different input representations and model combinations. We show that IndicNLP FastText embeddings perform better as compared to random initialization and original Facebook FastText embeddings when used with CNN, LSTM, and BiLSTM models. In fact, they perform competitively with more complex BERT based models. The difference in accuracies for using a static and trainable embedding layer is very small, where the trainable approach has a slight advantage in accuracy. The ULMFiT model performs better than basic CNN and LSTM based models. The IndicBERT has an upper hand over the mBERT model and all other smaller models utilizing randomly initialized embeddings. In the end, we highlight the fact the Marathi NLP has not received enough attention. There are only a couple of pre-trained models and relevant publicly available datasets for Marathi text classification. Both the datasets are based on the News domain.\\

\section*{Acknowledgments}
This work was done under the L3Cube Pune mentorship program. We would like to express our gratitude towards our mentors at L3Cube for their continuous support and encouragement.

\bibliography{main,acl2020}
\bibliographystyle{acl_natbib}

\end{document}